\documentclass[letterpaper]{article} 
\usepackage{aaai23}  
\usepackage{times}  
\usepackage{helvet}  
\usepackage{courier}  
\usepackage[hyphens]{url}  
\usepackage{graphicx} 
\usepackage{natbib}  
\usepackage{caption} 
\usepackage{algorithm}
\usepackage{algorithmic}

\usepackage{multirow}
\usepackage{subfigure}
\usepackage{amssymb}
\usepackage{amsmath}

\title{Quantized Feature Distillation for Network Quantization}
\author{
    Ke Zhu,
    Yin-Yin He,
    Jianxin Wu\thanks{J. Wu is the corresponding author.}
}
\affiliations{
    State Key Laboratory for Novel Software Technology, Nanjing University, China\\
    zhuk@lamda.nju.edu.cn, heyy@lamda.nju.edu.cn, wujx2001@nju.edu.cn
}

\begin{document}

\maketitle

\begin{abstract}
Neural network quantization aims to accelerate and trim full-precision neural network models by using low bit approximations. Methods adopting the quantization aware training (QAT) paradigm have recently seen a rapid growth, but are often conceptually complicated. This paper proposes a novel and highly effective QAT method, quantized feature distillation (QFD). QFD first trains a quantized (or binarized) representation as the teacher, then quantize the network using knowledge distillation (KD). Quantitative results show that QFD is more flexible and effective (i.e., quantization friendly) than previous quantization methods. QFD surpasses existing methods by a noticeable margin on not only image classification but also object detection, albeit being much simpler. Furthermore, QFD quantizes ViT and Swin-Transformer on MS-COCO detection and segmentation, which verifies its potential in real world deployment. To the best of our knowledge, this is the first time that vision transformers have been quantized in object detection and image segmentation tasks.
\end{abstract}

\section{Introduction}

Network quantization transfers a full precision (FP) network's weights and activations to their fixed point approximations without obvious accuracy drop. Recently, various approaches~\cite{QAT_EWGS,QAT_N2UQ,FQ-ViT} have been proposed and quantization aware training (QAT) becomes a mature paradigm for its ability to recover network accuracy even in extreme low bit settings.

Modern QAT methods are based on a general principle: optimizing quantization interval (parameters) with task loss~\cite{QAT_QIL}. Many variants have been put forward, such as non-uniform quantization~\cite{QAT_LCQ,QAT_N2UQ}, complex gradient approximation~\cite{QAT_DSQ,QAT_EWGS} or manually designed regularization~\cite{QAT_AUX,QAT_cluster_promote}. These methods are complex and \emph{simplicity} is often sacrificed. Another line of QAT methods introduce knowledge distillation~\cite{Hinton_KD} into the quantization phase. While the idea of quantization KD is straightforward (i.e., a full precision teacher helps recover the accuracy of a quantized student network), the implementation includes heuristic stochastic precision~\cite{QAT_KD_SPEQ}, manually defined auxiliary modules~\cite{QAT_AUX} or several stages~\cite{QAT_KD_QKD}. All these methods resort to logit distillation, which are then difficult to be applied in \emph{other computer vision tasks} (e.g., object detection).

\begin{figure}
    \centering
    \includegraphics[width=0.7\linewidth]{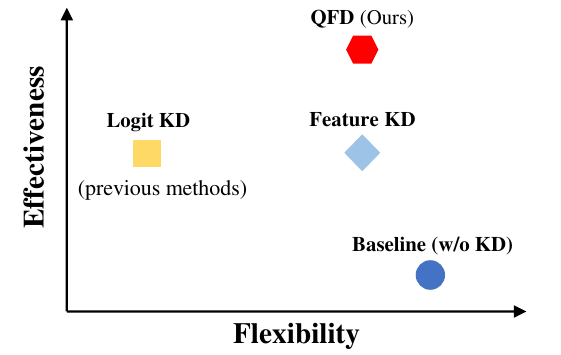}
    \caption{Illustration of different quantization KD methods. `Flexibility' stands for adaptability to different vision tasks, and `Effectiveness' indicates whether or not a method is friendly for network quantization (i.e., the ability to achieve high accuracy by the quantized network).}
    \label{fig:qualitative-analysis}
\end{figure}

\begin{figure}
	\centering
	\subfigure[compare with baseline]{
	\includegraphics[width=0.48\linewidth]{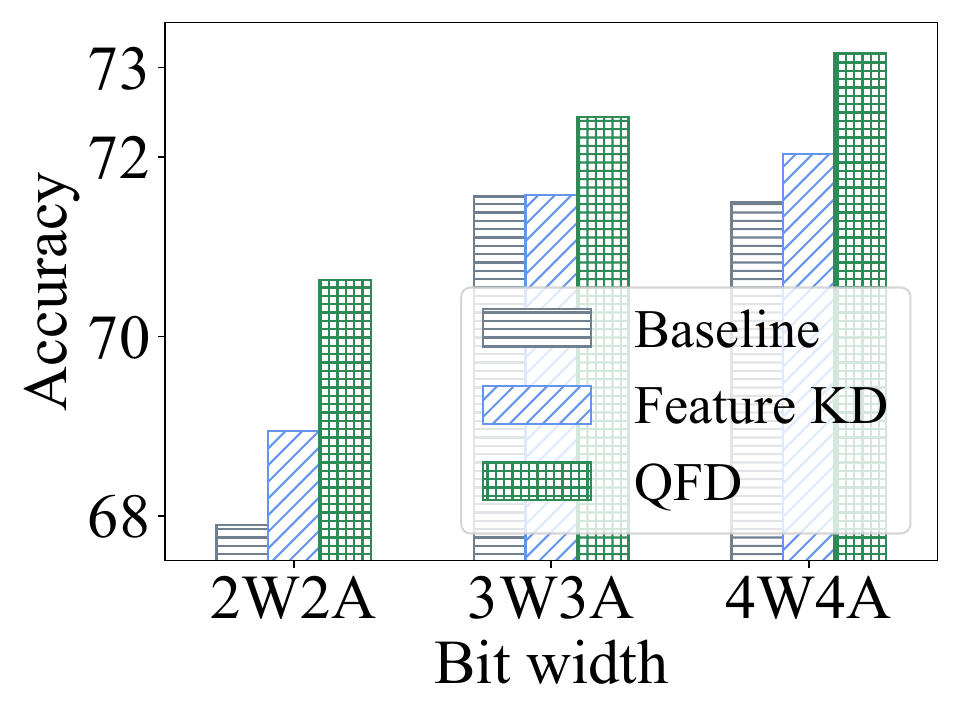}}
	\subfigure[compare with KD]{
	\includegraphics[width=0.48\linewidth]{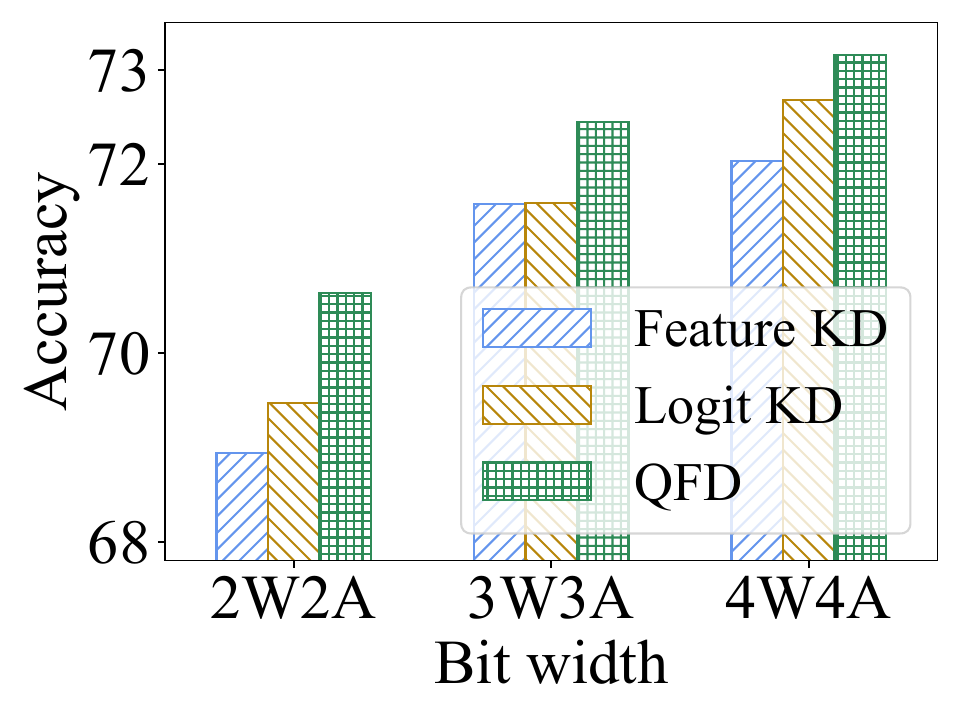}}
	\caption{The test accuracy of ResNet-18 models on CIFAR100 under 2-bit, 3-bit and 4-bit quantization settings. The left figure shows the results of the proposed QFD (with its teacher's feature binarized), float feature distillation and the baseline quantization, while the right compare three different KD methods (feature KD, logit KD and QFD). This figure is best viewed in color.}
	\label{fig:motivation_bar}
\end{figure}

In this paper, we propose a novel, simple, and effective distillation method targeting network quantization, \emph{quantized feature distillation} (QFD). Our motivation came from an important result in~\citet{ABC}: an \emph{FP model with only its output features binarized} can achieve similar or better accuracy compared with the full FP model. Then, it must be advantageous to use these quantized features as a teacher signal to help quantize the rest of the network (i.e., the student). On one hand, \emph{feature distillation is more flexible in comparison with logit distillation}, especially in tasks like object detection~\cite{Focal_det_distil}; on the other hand, it is natural to conjecture that \emph{it will be easier for the fully quantized student to mimic a fixed point feature representation rather than directly mimicking the floating-point logits or features}. In other words, the proposed QFD will be both more flexible and more effective (quantization friendly). This conjecture (and motivation for our QFD) is illustrated in Figure~\ref{fig:qualitative-analysis}, which is firmly supported by the experimental results in Figure~\ref{fig:motivation_bar} and Table~\ref{tab:differ-feature-bit}. 

These experiments that verified our motivation were carried out on CIFAR100 with ResNet-18 using 4 different quantization methods: the quantization baseline (a uniform quantizer without KD), logit distillation (logits as teacher signal), feature distillation (FP features as teacher signal) and our QFD (quantized features as teacher signal). All the 3 KD methods adopt the same quantizer as the baseline, and the teacher's feature is binarized (1-bit) in QFD. As shown in Figure~\ref{fig:motivation_bar}, QFD is not only superior than float feature distillation, but also surpasses logit distillation in all settings. The improvement of QFD over baseline is also consistent and large enough to recover the full precision model's accuracy (3-bit and 4-bit). We further quantize the teacher's feature to 4 bit and 8 bit using QFD. As shown in Table~\ref{tab:differ-feature-bit}, all feature distillation methods show improvement over the baseline. As teacher's bit width decreases, the accuracy of the teacher model drops, but the final distillation results have been consistently improved. These results show that our motivation is valid: \emph{quantized features are better teachers for network quantization!}

\begin{table}
	\centering
	\setlength{\tabcolsep}{3pt}
	\begin{tabular}{l|c|ccccc}
	\hline
		Method     & Teacher bit width & 1/1 & 2/2 & 3/3 & 4/4 \\
	  \hline
		 Baseline &  & 57.38 & 67.90 & 71.56 & 71.50  \\
		 \hline
	  \multirow{4}{*}{QFD} & 1 (71.14) &\textbf{62.62} & \textbf{70.63} & \textbf{72.45} & \textbf{73.16} \\
  
	  & 4 (71.30) & 61.60 & 69.63 & 71.60 & 72.18 \\
	  & 8 (71.53) & 61.45 & 69.98 & 71.43 & 72.28 \\
	  & 32 (71.88) & 59.66 & 68.94 & 71.58 & 72.03\\
	  \hline
	\end{tabular}
	\caption{ResNet-18 test accuracy on CIFAR-100. ``W/A'' means the weight and activation bit width in quantization. `Baseline' is the quantization without distillation. During QFD, the teacher's feature is quantized to 1, 4 and 8 bits with the teacher accuracy inside the brackets `()'. The 32-bit feature refers to full precision (normal feature distillation).}
	\label{tab:differ-feature-bit}
\end{table}

We tested QFD in not only classification but also detection and segmentation, and achieved state-of-the-art accuracy on diverse network structures (ResNet, MobileNet and ViT). Our contributions can be summarized as follows:
\begin{itemize}
    \item A novel quantization aware training KD method that is easy to implement.
    \item Remarkable accuracy advantage on classification, detection and segmentation benchmarks over previous quantization aware training methods.
    \item A first trial of quantizing the vision transformer structure on common object detection and segmentation tasks.
\end{itemize}

\begin{figure*}
  \centering
  \subfigure[Baseline]{
  \includegraphics[width=0.21\linewidth]{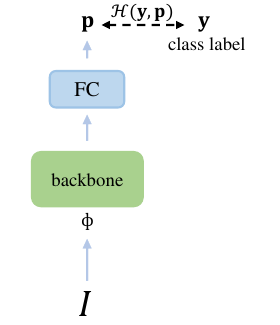}}
  \subfigure[Logit distillation]{
  \includegraphics[width=0.21\linewidth]{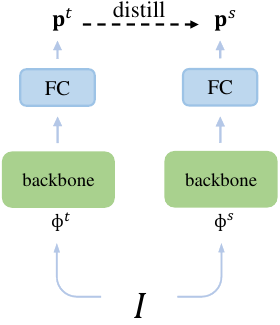}}
  \hspace{18pt}
  \subfigure[Feature distillation]{
  \includegraphics[width=0.21\linewidth]{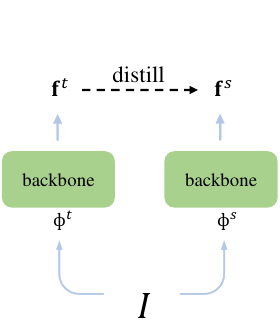}}
  \hspace{18pt}
  \subfigure[QFD (ours)]{
  \includegraphics[width=0.21\linewidth]{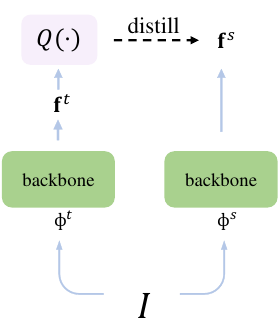}}
  \caption{The baseline quantization method and 3 different knowledge distillation quantization methods (logit distillation, feature distillation and our QFD). $I$ is the input image, $\phi$, $\mathbf{p}$ and $\mathbf{f}$ mean backbone, logits and feature, respectively. $t$ and $s$ denote the teacher and the student, respectively; $Q(\cdot)$ is the process of quantizing the teacher's features. The common cross entropy loss $\mathcal{H}(\mathbf{y},\mathbf{p})$ is calculated using the true label $\mathbf{y}$ and logits $\mathbf{p}$ as shown in (a). For clarity, we do not show the cross entropy loss $\mathcal{H}(\mathbf{y}, \mathbf{p}^{s})$ of all 3 distillation methods in (b), (c) and (d). This figure is best viewed in color.}
  \label{fig:method}
\end{figure*}

\section{Related Work}

Neural network quantization can be categorized into two paradigms: quantization aware training (QAT) and post training quantization (PTQ). We adopt QAT in this paper.  In this section, we will describe the basics, knowledge distillation, and vision transformers in QAT.

\textbf{Quantization aware training}. QAT~\cite{White_Paper} is a powerful paradigm to tackle low bit (e.g., 3- or 4-bit) quantization without significant accuracy drop. Integrating quantization operations into the computation graph is the key in QAT such that the weights and quantization parameters can be learned simultaneously through back-propagation. Early methods in this family focused on how to binarize models~\cite{BC,XNOR-Net}, fit quantizers with statistics~\cite{QAT_DoReFa,QAT_HWGQ,QAT_PACT}, or minimize local error~\cite{QAT_LQ-Net}, but they suffer from the incomplete or sub-optimal issue. Modern QAT methods adopt the principle of optimizing quantization interval with task loss~\cite{QAT_LSQ,QAT_QIL} and resort to much more complicated techniques, including non-uniform quantizer~\cite{QAT_LCQ,QAT_N2UQ}, gradient approximation~\cite{QAT_Distance_AQ,QAT_EWGS,QAT_DSQ} or extra regularization~\cite{QAT_bin_regular,QAT_cluster_promote,QAT_AUX}. Their simplicity and effectiveness, however, remain as big challenges.

\textbf{Knowledge distillation in quantization}. KD~\cite{Hinton_KD} is popular in various computer vision tasks~\cite{Dive,QAT_SPnet} and has gradually emerged in quantization aware training~\cite{QAT_KD_Apprentice,QAT_KD_modeldistill,QAT_KD_SPEQ,QAT_AUX}. The point of quantization KD is neat: utilizing a full-precision teacher to recover the accuracy of the quantized student network. However, recent methods are short of simplicity in that they involve complex stages~\cite{QAT_KD_QKD}, auxiliary modules~\cite{QAT_AUX} and dedicated mixed precision adjustment~\cite{QAT_KD_SPEQ}. Moreover, these methods all adopt logit distillation, which is in-flexible when KD is applied to object detection quantization~\cite{Det_distill}. Instead, we propose our quantized feature KD, which is both quantization friendly in terms of accuracy and flexible in terms of pipeline design.

\textbf{Quantizing vision transformers}. Vision Transformers (ViT) have boosted numerous vision tasks~\cite{ViT,Deit} and there is an urgent need to precisely quantize them so as to facilitate their practical usage~\cite{VAQF}. Recent methods~\cite{PTQ4ViT,QVIT-first,FQ-ViT} tried post training quantization~\cite{White_Paper} techniques to quantize ViT to 6- or 8-bit for only image classification. Low bit (3- or 4-bit) quantization and its applicability to detection and segmentation of ViT and variants remain unexplored. For the first time, we will answer both questions by exploring its quantization performance in all these settings and tasks.

\section{The Proposed Method}

We have revealed the motivation of the proposed QFD method in the introduction, and the results in Figure~\ref{fig:motivation_bar} and Table~\ref{tab:differ-feature-bit} not only supported this motivation but also initially verified the effectiveness of QFD. After introducing the preliminaries of neural network quantization (the QAT baseline we use), we will then move on to a detailed description of our proposed method.

\subsection{Preliminaries} \label{sec:pre}

We adopt the method in~\citet{QAT_EWGS} as our baseline method, which is a uniform quantizer composed of normalization, quantization and de-quantization steps. 

For any given full-precision data $v$ (a certain layer's weight or activation in the neural network), we define the quantization parameter $l$ and $u$, which represent the lower bound and upper bound of the quantization interval, respectively. The normalization step is as follow:
\begin{equation}
    \label{eq:clip}
    \hat{v} = clip\left(\frac{v-l}{u-l}, 0, 1\right) \,,
\end{equation}
where $clip(\cdot, min, max)$ clips data that lies outside the min-max range. Then, a quantization function is used,
\begin{equation}
    \label{eq:round}
    \widetilde{v} =  \frac{\lfloor(2^b - 1) \hat{v}\rceil}{2^b - 1} \,,
\end{equation}
in which $\lfloor \cdot \rceil$ is the rounding function and $b$ stands for the bit width of quantization. The operation $\lfloor(2^b - 1) \hat{v}\rceil$ maps $\hat{v}$ from the range $[0,1]$ to a discrete number in $\{0,1,...,2^b-1\}$. Finally, a de-quantization step is applied to output the quantized weight $\overline{v}_{W}$ or activation $\overline{v}_{A}$:
\begin{align}
    \overline{v}_{W} &= 2(\widetilde{v} - 0.5) \,, \text{ or,}\label{eq:dequant_w}\\
    \overline{v}_{A} &= \widetilde{v} \,,\label{eq:dequant_a}
\end{align}
where the quantized weight $\overline{v}_{W}$ is roughly symmetric about zero and the quantized activation $\overline{v}_{A}$ is positive considering the ReLU activation. Similar to~\citet{QAT_EWGS}, we use a trainable scale parameter $\alpha$ that is multiplied by the output quantized activation. 

During training, we adopt the straight through estimator (STE)~\cite{STE} to approximate the gradient of the rounding operator as 1:
\begin{equation}
    \frac{\partial \lfloor x \rceil}{x}  = 1 \,.
\end{equation}
The model's weights and these quantization parameters are learned simultaneously through back propagation.

\subsection{Quantized Feature Distillation}

We first define the basic notation in quantization aware training, then introduce our quantized feature distillation method, which is illustrated in Figure~\ref{fig:method}. For a given image $I$, it is first sent to a feature extractor $\phi(\cdot, \theta, \theta_{q})$ (the backbone, e.g., a CNN or ViT) to get a full precision feature vector $\textbf{f} \in \mathbb{R}^{D}$ (which is often obtained by a global average pooling),
\begin{equation}
\textbf{f} = \phi(I, \theta, \theta_{q}) \,,
\end{equation}
in which $D$ is the feature dimensionality, $\theta$ and $\theta_{q}$ represent the weight parameters and the quantization parameters of the model, respectively. $\textbf{f}$ is passed through a classifier to get the final logit $\textbf{p}\in \mathbb{R}^{C}$ with $C$ classes, which produces the cross entropy loss $\mathcal{H}(\textbf{y}, \textbf{p})$ along with the true class label $\textbf{y}\in \mathbb{R}^{C}$. The parameters $\theta$ and $\theta_q$ are learned through back propagation.

For our proposed QFD method, an image $I$ is separately sent to the teacher network $\phi^{t}(\cdot, \theta^{t})$ and the student network $\phi^{s}(\cdot, \theta^{s}, \theta^{s}_{q})$ to get features $\textbf{f}^{t}$ and $\textbf{f}^{s}$, respectively. The teacher's full precision feature $\textbf{f}^{t}$ will be quantized into lower bit (e.g., 1-bit or 4-bit) representation:
\begin{equation}
    \textbf{f}^{t} \stackrel{Q(\cdot)}{\longrightarrow} \overline{\textbf{f}^{t}}  \,,
\end{equation}
where the quantizer $Q(\cdot)$ is defined in Equations~(\ref{eq:clip})--(\ref{eq:dequant_a}). The feature quantizer $Q(\cdot)$ follows the activation quantization process described in the preliminaries.

The teacher's quantized feature then acts as the supervision signal to guide quantization of the student network by the mean squared loss $\mathcal{L}(\cdot,\cdot)$, and the student still produces its usual cross entropy loss $\mathcal{H}(\cdot,\cdot)$ with the true label $\textbf{y}$. The overall optimization objective is:
\begin{equation}
\label{eq:loss_func}
    \mathop{\arg\min}_{\theta^{s}, \theta^{s}_{q}} \lambda \mathcal{L}(\textbf{f}^{s}, \overline{\textbf{f}^{t}}) + (1-\lambda) \mathcal{H}(\textbf{y}, \textbf{p}^s) \,.
\end{equation}
Here $\lambda$ is used to weigh the importance of the distillation loss with respect to the cross entropy loss. For simplicity, we set $\lambda=0.5$ except in ablation studies.

\section{Experimental Results}

In this section, we will first describe the general experimental settings, then present the results of our QFD on classification, object detection and segmentation benchmarks.

\subsection{Experimental Settings}

During training, we first take a few epochs (roughly 1/10 of the total number of training epochs) to quantize the teacher's feature to fixed low bit (e.g., 2-bit) before starting our QFD training. Following previous QAT works~\cite{QAT_AUX,QAT_EWGS}, we conduct our experiments on the CIFAR, ImageNet, CUB and MS-COCO datasets. All experiments use PyTorch~\cite{pytorch} with 8 GeForce RTX 3090. The evaluation metrics for classification and detection (segmentation) are top-1 (top-5 is also used on ImageNet) accuracy and AP (average precision), respectively.

\textbf{Classification settings}. We experiment with ResNet-20 on CIFAR10 and ResNet-18/32 on CIFAR100. On both CIFAR datasets~\cite{CIFAR100}, we use SGD with learning rate of 0.004, weight decay of 0.0005 and train 200 epochs in total. The input resolution is 32$\times$32, and random flip and random crop are used as data augmentation. On ImageNet~\cite{ImageNet}, we train ResNet-18, ResNet-34 and MobileNet-v2 for 100 epochs. The initial learning rate and the momentum is 0.01 and 0.9, respectively. The weight decay is set to 1e-4, 5e-5 and 2.5e-5 for 4-bit, 3-bit and 2-bit, respectively, following~\citet{QAT_bin_regular,QAT_LSQ}. We adopt random resized crop and random flip as data augmentation and set input resolution as $224\times224$. On CUB200~\cite{CUB200}, resolution and augmentation are the same as those on ImageNet. We train ViT Small, ViT Base, Deit Small, Deit Base, Deit Tiny for 90 epochs with batch size 64, following a cosine scheduler. The learning rate and weight decay are 5e-3 and 5e-4, respectively. We take 3 runs for CIFAR and CUB since these results have larger variance.

\textbf{Object detection and segmentation settings}. We train RetinaNet~\cite{Retinanet} detectors with ResNet as backbones, and explore vision transformer detection and segmentation quantization using ViT and Swin Transformer~\cite{SwinTransformer} pretrained with the newly proposed self-supervised method MAE~\cite{SSL-MAE}. The object detector structure follows~\citet{ViT-Det}. For RetinaNet, we train 90k iteration with a base learning rate 0.001. Following~\citet{QAT_LCQ}, we quantize all the layers (backbone, FPN and detection head) except the input and the output of the whole network, and utilize BN after FPN and detection head. To implement our QFD method, we quantize the teacher's FPN layer output to 8-bit and then use this `quantized feature' for distillation, following the concept of previous object detection distillation works~\cite{Det_distill}. For ViT and Swin Transformer, we quantize all the linear layer in their backbone and evaluate them on detection and segmentation tasks. All these models take 2 runs on MS-COCO and are implemented with Detectron2~\cite{detectron2}.

\begin{table}
	\centering
	\small
	\begin{tabular}{l|l|c|ccc}
	\hline
	  Model     & Methods     & 32/32 & 2/2 & 3/3 & 4/4 \\
	  \hline
	  \multirow{7}{*}{R20}
  
	  & LQ-Net & 92.1 & 90.20 & 91.60 & - \\
	  & SPEQ & 92.1  & 91.40 & - & - \\
	  & APRT$^*$ & 91.6 & 88.60 & 91.80 & 92.20 \\
	  \cline{2-6}
	  & Baseline & \multirow{4}{*}{92.1} & 90.49 & 91.74 & 92.09\\
	  & Feature KD &  & 90.84 & 91.98 & 91.12 \\
	  & Logit KD &  & 90.61 & 91.78 & 92.02 \\
	  & QFD &  & \textbf{91.41} & \textbf{92.64} & \textbf{93.07}   \\
	  \hline
	\end{tabular}
	\caption{ResNet-20 top-1 accuracy on CIFAR10. ``W/A'' represent the bit-width of weight and activation. SPEQ~\cite{QAT_KD_SPEQ} and APRT$^*$~\cite{QAT_KD_Apprentice} are two logit distillation methods.}
	\label{tab:R20-CIFAR10}
\end{table}

\begin{table}
	\centering
	\small
	\setlength{\tabcolsep}{4.6pt}
	\begin{tabular}{l|l|c|cccc}
	\hline
	  Model     & Methods     & 32/32 & 1/1 & 2/2 & 3/3 & 4/4 \\
	  \hline
	  \multirow{5}{*}{R18} & Auxi &  -    & -  & 67.90 &- &- \\
	  & Baseline & \multirow{4}{*}{71.9} & 57.38 & 67.90 & 71.56 & 71.50  \\
	  
	  & Feature KD &  & 59.66 & 68.94 & 71.58 & 72.03 \\
	  & Logit KD&  & 62.60 & 69.47 & 71.59 & 72.68 \\
	  & QFD &  & \textbf{62.62} & \textbf{70.63} & \textbf{72.45} & \textbf{73.16} \\
	  \hline
	  \multirow{5}{*}{R32} & APRT$^*$ & 70.8 & - & 63.50 & 70.30 & 71.50 \\
  
	  & Baseline &  \multirow{4}{*}{70.5}  & 56.09 & 67.38 & 69.24 & 70.13 \\
	  & Feature KD  & & 54.60 & 67.08 & 69.43 & 70.23 \\
	  & Logit KD & & 55.46 & 67.72 & 70.25 & 71.14 \\
	  & QFD &  & \textbf{56.84} & \textbf{68.30} & \textbf{70.65} & \textbf{71.52}  \\
	  \hline
	\end{tabular}
	\caption{ResNet-18/32 on CIFAR-100, ``W/A'' represents the bit width of weight and activation. Both Auxi~\cite{QAT_AUX} (a variant of KD) and APRT$^*$ utilize logit distillation.}
	\label{tab:CIFAR100}
\end{table}

\begin{table*}
	\centering
	\small
	\begin{tabular}{l|l|c|ccc}
	\hline
	  Architecture     & Methods     & 32/32  & 2/2 & 3/3 & 4/4 \\
	  \hline
	  \multirow{8}{*}{ResNet-18} & Auxi & -  & 66.7/87.0     \\
	  & SPEQ~\cite{QAT_KD_SPEQ} & 70.3 & 67.4     \\
	  & QIL~\cite{QAT_QIL} & 70.2 & 65.7 & 69.2 &  70.1 \\
	  & LSQ + BR\cite{QAT_bin_regular} & 70.5 & 67.2/87.3 & 69.9/89.1 & 70.8/89.6 \\
	  
	  & EWGS~\cite{QAT_EWGS} & 69.9  & 67.0 & 69.7 & 70.6 \\
	  & DAQ~\cite{QAT_Distance_AQ} & 69.9  & 66.9 & 69.6 & 70.5 \\
	  & LSQ+ \cite{LQ+} & 70.1 & 66.7 & 69.4 & 70.7 \\
	  & QFD  & 70.5 & \textbf{67.6}/\textbf{87.8}  &\textbf{70.3/89.4}  & \textbf{71.1}/\textbf{89.8}     \\
	  \hline
	   \multirow{7}{*}{ResNet-34} & Auxi~\cite{QAT_AUX}+DoReFa & - & 71.2/89.8    \\
	  & SPEQ~\cite{QAT_KD_SPEQ} & 73.6  & 71.5    \\
	  & QKD~\cite{QAT_KD_QKD}(teacher R50) & 73.5 & 71.6/90.3 & 73.9/91.4 & 74.6/92.1\\
	  & QIL~\cite{QAT_QIL} & 73.7  & 70.6 & 73.1 &  73.7 \\
  
	  & DAQ~\cite{QAT_Distance_AQ} & 73.3  & 71.0 & 73.1 & 73.7 \\
  
	  & EWGS~\cite{QAT_EWGS} & 73.3  & 71.4 & 73.3 & 73.9 \\
	  & QFD  & 73.3  & \textbf{71.7}/\textbf{90.4} & \textbf{73.9}/\textbf{91.7} & \textbf{74.7}/\textbf{92.3}   \\
	  \hline
	  \multirow{4}{*}{MobileNetV2}    &  QKD~\cite{QAT_KD_QKD}& \multirow{4}{*}{71.8}  & 45.7/68.1 & 62.6/84.0 & 67.4/87.0 \\
	  & LSQ+KURE~\cite{QAT_bin_regular} &   & 37.0/62.0 & 65.9/86.8 & 69.7/89.2 \\
	  & LSQ$^*$~\cite{QAT_LSQ} &  & 46.7/71.4 & 65.3/86.3 & 69.5/89.2  \\ 
	  & QFD &   & \textbf{52.8}/\textbf{77.1} & \textbf{66.4}/\textbf{87.0} & \textbf{70.5}/\textbf{89.5} \\
	  \hline
	\end{tabular}
	\caption{Comparing with state-of-the-art methods on ImageNet. SPEQ, QKD and Auxi all adopt KD in their quantization training. ``W/A'' in the first row represents the bit width of weights and activations. For the proposed QFD method, we report the top-1 (\%) and top-5 (\%) accuracy of each result item with `/' to separate them.}
	\label{tab:ImageNet}
\end{table*}

\subsection{Classification Results}

\textbf{CIFAR10/100}. We first validate our proposed QFD method using ResNet models on CIFAR10 and CIFAR100, each containing 50,000 training images and 10,000 validation images, while the latter serves as a finer categorization (100 classes) than the former (10 classes).

For ResNet-20 models on CIFAR10 (results shown in Table~\ref{tab:R20-CIFAR10}), we run the baseline, feature knowledge distillation (`Feature KD'), logit distillation (`Logit KD') and our proposed quantized feature distillation method (`QFD'). We quantize ResNet-20 to 2-bit, 3-bit and 4-bit for both weights and activations (``W/A''). Following previous work~\cite{QAT_LQ-Net,QAT_KD_SPEQ}, we quantize all the layers except the input to the backbone and the last fully connected layer (i.e., the classifier). SPEQ~\cite{QAT_KD_SPEQ} and APRT$^*$ both utilize the logit distillation method, and LQ-Net~\cite{QAT_LQ-Net} is a quantization aware training method. As can be seen in Table~\ref{tab:R20-CIFAR10}, our QFD surpasses previous knowledge distillation quantization methods SPEQ and APRT$^*$ by a large margin, and is better than both Feature KD and Logit KD. Note that our QFD has achieved even higher accuracy than the full-precision model under 3-bit (92.64\%) and 4-bit (93.07\%) settings.

We also validate our QFD on CIFAR100 using ResNet-18 and ResNet-32. Similar to the experiments on CIFAR10, we reproduce the baseline, feature distillation and logit distillation methods. As shown in Table~\ref{tab:CIFAR100}, Feature KD and Logit KD are generally better than the Baseline, showing the power of knowledge distillation. Our QFD is better than all of them, especially in extreme low bit scenarios (1-bit and 2-bit). Our method can almost recover the accuracy of a full precision model in 2, 3, 4 bit for ResNet-18 and ResNet-32. Especially for 1-bit ResNet-32, only our QFD shows improvement over the Baseline (56.84\% vs 56.09\%).

\textbf{ImageNet results}. We compare the proposed QFD with other QAT methods on the ImageNet1k dataset. The results can be found in Table~\ref{tab:ImageNet}. The proposed quantization feature distillation surpasses previous methods (including other knowledge distillation methods SPEQ, QKD and Auxi) with ResNet-18, ResNet-34 and MobileNetV2 models under different bit settings. Note that Auxi uses a manually designed auxiliary module, SPEQ needs empirical exploration of stochastic precision, and QKD involves larger teacher models (e.g., ResNet-50 to distill ResNet-34). In comparison, our method is both \emph{simpler} in concept and \emph{more effective} in accuracy, especially for MobileNetV2, where our QFD surpasses QKD by a large margin (increase by 7.1\%, 3.8\% and 3.1\% under 2-bit, 3-bit and 4-bit quantization settings, respectively). For ResNet series models, our QFD perfectly recovers the full precision's top-1 accuracy under 3- and 4-bit quantization (4-bit ResNet-34 with 74.7\% top-1 even surpasses its full precision counterpart by 1.3\%). Meanwhile, the accuracy of MobileNetV2 is relatively more difficult to recover under low bit, possibly due to its large channel variance, as pointed out by~\citet{White_Paper}. But our QFD is still better than other methods.

\textbf{CUB200 with ViT}. We also quantize vision transformers on the image classification benchmark CUB200~\cite{CUB200}, which contains 200 categories of birds, with 5,994 and 5,794 images for training and testing, respectively. Specifically, we quantize the linear layer in multi-layer-perceptron (MLP) and multi-head-attention (MHA) to 3- or 4-bit, using different structures of ViT~\cite{ViT} and Deit~\cite{Deit}, including ViT Small, ViT Base, Deit Tiny, Deit Small and Deit Base. We also list the accuracy of the teacher network with quantized features (a preprocessing step of our QFD method). As Table~\ref{tab:cub200} shows, although quantizing only the feature brings a slight accuracy drop to the original FP model, the improvement of QFD method over Baseline is significant and consistent. But, there is still a gap between 4-bit and the FP models. Quantizing transformer still remains a challenging task.

\begin{table*}
	\centering
	\small
	\begin{tabular}{l|c|ccccc}
	\hline
			Methods     & Bit (W/A) & Vit Small & ViT Base & Deit Tiny & Deit Small &  Deit Base\\
	  \hline
	   Full precision & 32/32 & 82.64 & 89.44 & 73.58 & 81.29 & 84.14  \\
	   Feature 4-bit & 32/32 & 82.19 & 88.89 & 72.04 & 79.84 & 83.95 \\
	   Feature 8-bit & 32/32  & 82.21 & 88.71 & 72.28 & 80.00 & 83.70 \\
	  \hline
	   Baseline & 3/3 & 74.40 & 83.15 & 61.11 & 64.77  & 69.24 \\
	   QFD & 3/3 & \textbf{77.01} & \textbf{84.28} & \textbf{62.24} &\textbf{67.05}  & \textbf{70.68} \\
	  \hline
	   Baseline & 4/4 & 78.44 & 86.74 & 69.36 & 74.56 & 78.68 \\
	   QFD & 4/4  & \textbf{81.15} & \textbf{87.42} & \textbf{69.68} & \textbf{76.44} & \textbf{81.67} \\
	  \hline
	\end{tabular}
	\caption{Vision transformers on CUB200. ``Feature $k$-bit'' means the performance of full precision (FP) teacher with its feature quantized to $k$-bit. Here we utilize the teacher with $4$-bit feature (``Feature 4-bit'') to distill the student network.}
	\label{tab:cub200}
\end{table*}

\subsection{Object Detection Results}

\textbf{RetinaNet}. RetinaNet~\cite{Retinanet} is a one-stage object detector composed of the backbone, FPN~\cite{FPN} and detection head. On the MS-COCO dataset~\cite{MS-COCO}, we quantize all its layers to 4-bit and 3-bit using the proposed QFD method (including the convolution operation in the skip connection) except the input to the backbone and the output in the detection head. Following previous work~\cite{QAT_LCQ,QAT_AUX}, our quantization training is finetuned using the full precision model (`FP' in Table~\ref{tab:retinanet-4bit}).

\begin{table}
	  \small
	  \centering
	  \setlength{\tabcolsep}{4pt}
	  \begin{tabular}{l|l|cccccc}
	  \hline
		Arch     & Methods &  AP & AP$_{50}$ & AP$_{75}$ & AP$_{S}$ & AP$_{M}$ & AP$_{L}$\\
		\hline
		\multirow{4}{*}{R18} & FP & 33.4 & 52.3 & 35.7 & 19.3 & 36.2 & 44.0 \\
		& Auxi & 31.9 & 50.4 & 33.7 & 16.5 & 34.6 & 42.3 \\
		& LCQ & 32.7 & 51.7 & 34.2 & 18.6  & 35.2 & 42.3 \\
		& QFD & \textbf{33.7} & \textbf{52.4}	& \textbf{35.6} & \textbf{19.7} & \textbf{36.3} & \textbf{44.5} \\
		\hline
		\multirow{4}{*}{R34} & FP & 37.1 & 56.7 & 39.6 & 22.7 & 41.0 & 47.6 \\
		& Auxi & 34.7 & 53.7 & 36.9 & 19.3 & 38.0 & 45.9 \\
		& LCQ & 36.4 & 55.9 & 38.7 & 21.2 & 40.0 & 46.6 \\
		& Ours & \textbf{37.0} & \textbf{56.4} & \textbf{39.4} & \textbf{22.8} & \textbf{40.5} & \textbf{48.1}\\
		\hline
		\multirow{4}{*}{R50} & FP & 38.6 & 58.3 & 41.5 & 24.1 & 42.2 & 49.7 \\
		& Auxi & 36.1 & 55.8 & 38.9 & 21.2 & 39.9 & 46.3 \\
		& LCQ & 37.1 & 57.0 & 39.6 & 21.2 & 40.8 & 47.1 \\
	
		& QFD & \textbf{38.2} & \textbf{57.5} &	\textbf{40.9} & \textbf{23.1}& \textbf{41.4} & \textbf{49.3} \\
		\hline
	  \end{tabular}
      \caption{RetinaNet detector with ResNet-18, 34 and 50 backbones under 4-bit setting on MS-COCO. Note that LCQ~\cite{QAT_LCQ} adopt a more complex quantizer (non-uniform quantizer). `FP' stands for full precision.}
	  \label{tab:retinanet-4bit}
\end{table}

For the teacher network, we first quantize its output feature at the $p3$ level to 8-bit since it contains the most gradient flow in the FPN graph~\cite{Retinanet}, then use it as the quantized feature to distill a student RetinaNet. Empirically, we find that utilizing the quantized feature of all the FPN level (including $p3$, $p4$, $p5$, $p6$, $p7$), the common approach in object detection distillation~\cite{Det_distill}, achieves similar accuracy but is unstable. For simplicity, we only use $p3$ for feature distillation and do \emph{not} involve any complex operation like distinguishing foreground and background features~\cite{Focal_det_distil,Det_distill}. The quantized feature distillation loss occupy about 1/5 of the total detection loss, and the RetinaNet structure strictly follows previous quantization work~\cite{QAT_LCQ}.

Table~\ref{tab:retinanet-4bit} shows the result of quantizing RetinaNet to 4-bit. Our QFD (ResNet18/34/50 as backbone) surpasses previous methods by a large margin. Especially for ResNet-18, our QFD even surpasses its full precision counterpart (improvement of 0.3\%, 0.4\% and 0.5\% for $AP$, $AP_{S}$ and $AP_{L}$, respectively). Our accuracy drop from the full precision ones with ResNet-34 is negligible as well, with a slight decrease of 0.1\% on $AP$ and 0.2\% on $AP_{75}$.
	
Table~\ref{tab:retinanet-3bit} shows the result of quantizing RetinaNet to 3-bit. Unlike 4-bit quantization, 3-bit is more challenging and difficult to optimize due to its limited representation power. Empirically we find ResNet-34 often face unstable training issue, such that we elongate its warmup iterations while keeping the total training iterations fixed. Overall, our QFD outperforms previous state-of-the-art method by a relatively large margin, especially for ResNet-18 where the improvement of $AP_{M}$ and $AP_{L}$ over LCQ~\cite{QAT_LCQ} is 0.7\% and 2.2\%, respectively.

\begin{table}
	  \centering
	  \small
	  \setlength{\tabcolsep}{4pt}
	  \begin{tabular}{l|l|cccccc}
	  \hline
		Arch     & Methods &  AP & AP$_{50}$ & AP$_{75}$ & AP$_{S}$ & AP$_{M}$ & AP$_{L}$\\
		\hline
		\multirow{4}{*}{R18} & FP & 33.4 & 52.3 & 35.7 & 19.3 & 36.2 & 44.0 \\
		& APoT & 31.2 & 50.1 & 32.8 & 18.0 & 33.5 & 40.6 \\
		& LCQ & 31.3 & 50.2 & 33.1 & 17.6 & 33.8 & 40.4 \\
		& QFD & \textbf{32.0} & \textbf{50.3} & \textbf{33.9} & \textbf{18.0} & \textbf{34.5} & \textbf{42.6} \\
		\hline
		\multirow{3}{*}{R34} & FP & 37.1 & 56.7 & 39.6 & 22.7 & 41.0 & 47.6 \\
		& APoT & 35.2 & 54.9 & 37.1 & 19.7 & \textbf{39.1} & 45.3 \\
		& QFD & \textbf{35.6} & \textbf{54.9} & \textbf{37.9} & \textbf{21.6} & \underline{39.0} & \textbf{45.6} \\
		\hline
		\multirow{3}{*}{R50} & FP & 38.6 & 58.3 & 41.5 & 24.1 & 42.2 & 49.7 \\
		& LCQ & 36.1 & \textbf{56.2} & 38.4 & 21.7 & 
		\textbf{39.9} &  46.1 \\
		& QFD & \textbf{36.5} & \underline{56.1} & \textbf{39.0} & \textbf{22.1} & \underline{39.5} & \textbf{47.2}  \\
		\hline
	  \end{tabular}
	  \caption{RetinaNet detector with ResNet-18, 34 and 50 backbones under 3-bit setting on MS-COCO. APoT~\cite{QAT_APoT} and LCQ both use non-uniform quantizers while our QFD uses the simple uniform one.}
	  \label{tab:retinanet-3bit}
\end{table}

\textbf{The ViT structure}. Lastly, we explore quantizing ViT. To the best of our knowledge, this is the first time that ViT has been quantized in detection and segmentation tasks. We tried ViT~\cite{ViT} and Swin Transformer~\cite{SwinTransformer} pretrained on ImageNet1k and ImageNet21k, respectively, using the self-supervised learning methods MAE~\cite{SSL-MAE}. The detection pipeline follows the newly published ViTDet~\cite{ViT-Det}. Since most parameters are in the linear layer of MLP and MHA in the backbone transformer blocks, we quantize only the linear layers in the backbone and run the baseline quantization (without QFD distillation) under 8, 6, 4 bit settings.

As shown in Table~\ref{tab:vit-det}, 8-bit or 6-bit quantization for linear layer of ViT and Swin Transformer (`SwinB' means the base structure of it) is roughly enough to recover its detection and segmentation accuracy, demonstrating the potential to deploy vision transformers on real-world hardware devices~\cite{MQbench}. On the contrary, quantizing vision transformers to 4-bit leads to noticeable performance drop, possibly due to its limited representation capacity. We further analyze the impact of MHA and MLP by alternatively quantizing each only, and the results in Table~\ref{tab:vit-det} convey an interesting observation: quantizing vision transformer in detection and segmentation is \emph{not at all sensitive to the attention layer, but to the linear layers in the MLP}. Note that the performance of 4/4$^{a}$ in ViT and SwinB even surpassed its 8/8 counterpart. It is possible because the MLP layer is seriously affected by the inter-channel variation in LayerNorm inputs~\cite{FQ-ViT,LayerNorm}, while the MHA layer contains additional operations which might mitigate this effect.

\begin{table}
	  \centering
	  \small
	  \setlength{\tabcolsep}{3pt}
	  \begin{tabular}{l|c|cccccc}
	  \hline
		\multirow{2}{*}{Arch}     & \multirow{2}{*}{Bit} & \multicolumn{3}{c}{Detection} & \multicolumn{3}{c}{Segmentation} \\
		\cline{3-8}
		& & AP & AP$_{50}$ & AP$_{75}$ & AP & AP$_{50}$ & AP$_{75}$\\
		\hline
		\multirow{6}{*}{ViTDet} & FP & 50.8 & 71.6 & 56.3 & 45.3 & 69.2 & 49.3\\
		& 8/8 & 50.1 & 70.9 & 54.9 & 44.5 & 68.2 & 48.3 \\
		& 6/6 & 49.8 & 70.7 & 54.7 & 44.3 & 68.0 & 48.2 \\
		& 4/4 & 46.7 & 67.5 & 51.0 & 41.4 & 64.4 & 44.4 \\
		& 4/4$^{a}$ & \textbf{50.3} & \textbf{71.0} & \textbf{55.2} & \textbf{44.5} & \textbf{68.3} & \textbf{48.1}\\
		& 4/4$^{m}$ & 47.6 & 68.4 & 52.1 & 42.3 & 65.6 & 45.5\\
		\hline
		\multirow{6}{*}{SwinB} & FP & 53.6 & 72.6 & 58.4 & 46.1 & 70.1 & 50.2 \\
		& 8/8 & 52.9 & 71.8 & 57.9 & 45.5 & 69.0 & 49.6 \\
		& 6/6 & 52.7 & 71.7 & 57.3 & 45.4 & 68.9 & 49.5 \\
		& 4/4 & 51.6 & 70.5 & 56.2 & 44.4 & 67.8 & 47.9\\
		& 4/4$^a$ & \textbf{53.2} & \textbf{72.1} & \textbf{58.1} & \textbf{45.7} & \textbf{69.5} & \textbf{49.6} \\
		& 4/4$^m$ & 52.3 & 71.2 & 57.1 & 45.0 & 68.4 & 49.1 \\
		\hline
	  \end{tabular}
	  \caption{Results of ViT and Swin Transformer quantization on MS-COCO detection and segmentation tasks. 4/4$^a$ means quantizing multi-head-attention (MHA) layer only, while 4/4$^m$ quantizes MLP layer only. All the other bit settings quantize both MHA and MLP.}
	  \label{tab:vit-det}
\end{table}

\subsection{Ablation Studies}

\textbf{Effect of $\lambda$ \& Quantizing feature}. We verify the only hyper-parameters $\lambda$ defined in Eq.~\ref{eq:loss_func} on ImageNet and CUB under different bit settings (2-bit, 3-bit and 4-bit). The result can be found in Tables~\ref{tab:lambda-mobile} to~\ref{tab:lambda-resnet34}. For both CNNs and vision transformers on various datasets, all the $\lambda$ values lead to improvements over the baseline quantization method and our method is not sensitive to the value of $\lambda$. Interestingly, in Table~\ref{tab:lambda-resnet34} (4-bit ResNet-34 quantization on ImageNet), our QFD further increases the accuracy by 1.1\% top-1 accuracy even when the baseline quantization method has already surpassed its full precision counterpart. Hence, we can choose $\lambda=0.5$ by default for simplicity.

\begin{table}
  \centering
  \small
  \begin{tabular}{cccccc}
  \hline
    $\lambda$     & bit  & top-1 acc & top-5 acc  \\
  \hline
    Full-precision & 32/32 & 71.8 & 90.2 \\
    Baseline  & 2/2 & 47.1 & 72.1 \\
    0.1 & 2/2 & 50.8 & 75.4 \\
    0.3 & 2/2 & \textbf{52.8} & \textbf{77.1} \\
    0.5 & 2/2 & 49.1 & 74.5 \\
    \hline
  \end{tabular}
  \caption{Results of different distillation hyper-parameter $\lambda$ of 2-bit MobileNetV2 on ImageNet. `acc' means accuracy.}
  \label{tab:lambda-mobile}
\end{table}

\begin{table}
  \centering
  \small
  \begin{tabular}{cccccc}
  \hline
    $\lambda$     & bit  & best acc & last acc  \\
  \hline
    Full-precision & 32/32 & 82.64 & 82.38 \\
    Baseline  & 3/3 & 74.40 & 74.40 \\
    0.1 & 3/3 & 75.66 & 74.78 \\
    0.5 & 3/3 & \textbf{77.01} & \textbf{76.77} \\
    0.9 & 3/3 & 76.39 & 76.08 \\
    \hline
  \end{tabular}
  \caption{Results of different distillation hyper-parameter $\lambda$ with 3-bit ViT Small on CUB200. `acc' stands for accuracy.}
  \label{tab:lambda-vit}
\end{table}

\begin{table}
  \centering
  \small
  \begin{tabular}{cccccc}
  \hline
    $\lambda$     & bit  & top-1 & top-5  \\
  \hline
    Full-precision & 32/32 & 73.3 & 91.4 \\
    Baseline  & 4/4 & 73.6 & 91.5 \\
    0.1 & 4/4 & 73.9  & 91.6  \\
    0.5 & 4/4 & \textbf{74.7} & \textbf{92.3} \\
    0.7 & 4/4 & 74.2 & 91.9 \\
    \hline
  \end{tabular}
  \caption{Results of different distillation hyper-parameter $\lambda$ of 4-bit ResNet34 on ImageNet. `acc' stands for accuracy.}
  \label{tab:lambda-resnet34}
\end{table}

\textbf{Effect of quantizing the teacher's feature}.  Meanwhile, we show the accuracy of the teacher network (with its feature quantized). As shown in Table~\ref{tab:quantizing-feature}, the feature-quantized teacher networks almost have no difference with the original in terms of accuracy. Hence, the accuracy improvement is brought by QFD, not because the teacher's accuracy is higher than that of the baseline.

\begin{table}
	\small
	\centering
	\begin{tabular}{ccc}
	\hline
	            & Full precision acc  & Teacher acc  \\
	\hline
	  ResNet-18 & 70.5 & 70.9\\
	  ResNet-34 & 73.3 & 73.3\\
	  MobileNetV2 & 71.8 & 71.1\\
	  \hline
	\end{tabular}
	\caption{The teacher's accuracy with its feature quantized to 4-bit on ImageNet classification. Quantizing feature to low bit has similar accuracy as that of the original FP model.}
	\label{tab:quantizing-feature}
\end{table}

\begin{figure}
	\centering
   \subfigure[ResNet-50 backbone]{
	\includegraphics[width=0.45\linewidth]{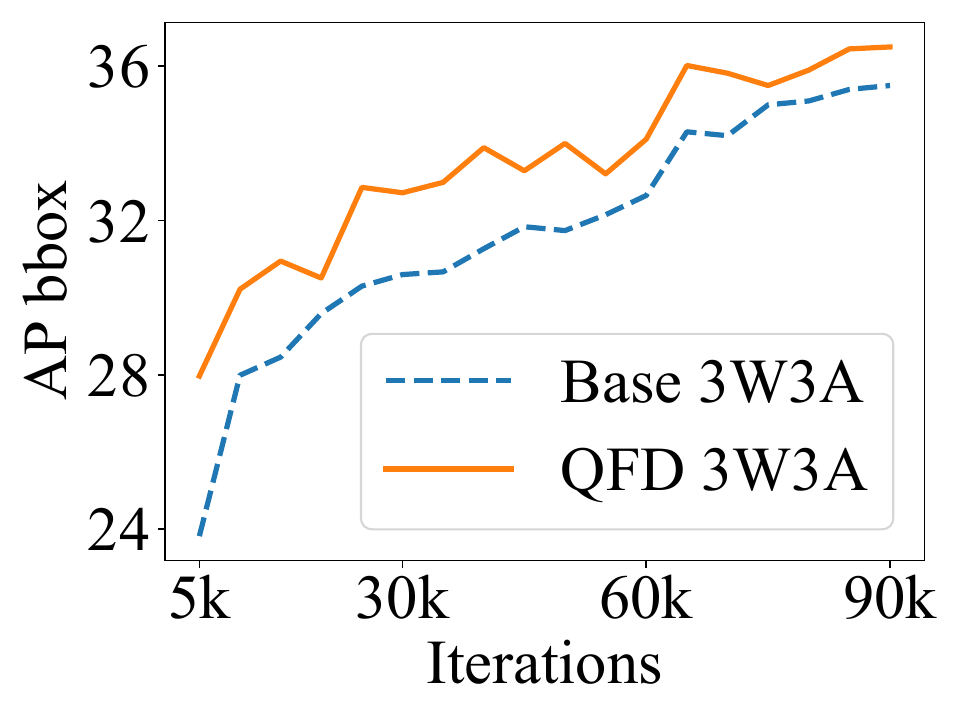}}
	\subfigure[ResNet-18 backbone]{
	 \includegraphics[width=0.45\linewidth]{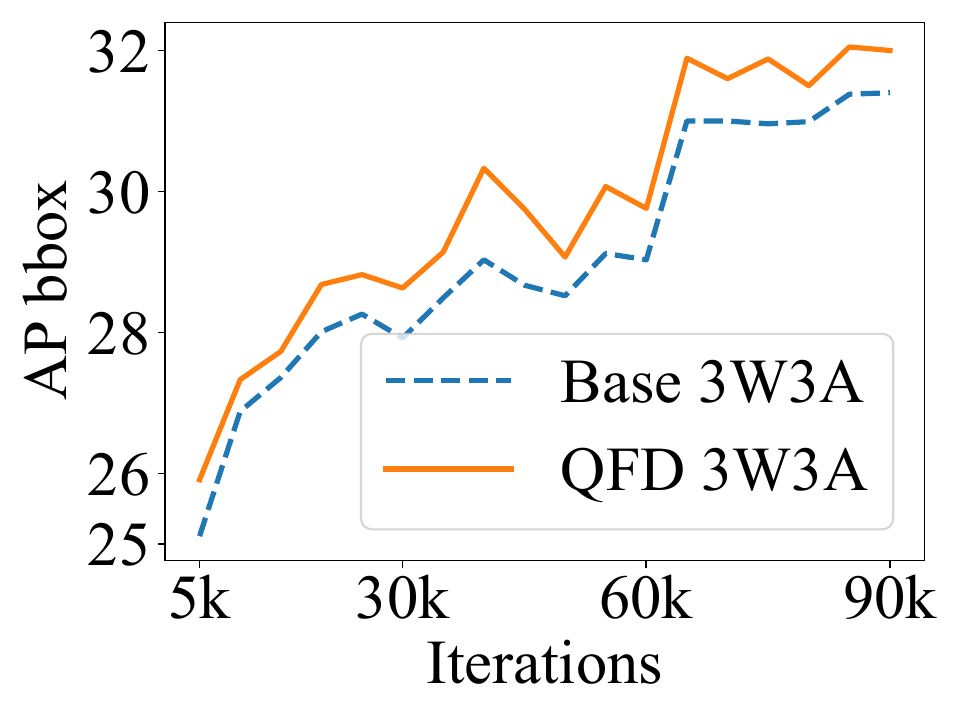}
	}
	\caption{Comparison of QFD with baseline using 3 bit RetinaNet (ResNet backbone) on MS-COCO detection tasks.}
	\label{fig:iteration-coco}
  \end{figure}

\textbf{Consistent improvement of detection}. Finally in this section, we plot the convergence curve of RetinaNet using either the baseline quantization or our QFD distillation on MS-COCO detection tasks. The results using ResNet-18 and ResNet-50 backbones under 3-bit quantization can be found in Figure~\ref{fig:iteration-coco}. There is no doubt that QFD makes consistent improvement over the baseline throughout the whole training process, demonstrating the generality of our methods: it is not only suitable for classification, but also boosts object detection performance as well.

\section{Conclusions}

In this paper, we proposed a novel and easy to implement feature distillation method QFD in quantization aware training. We first qualitatively illustrated QFD's advantages: simple, quantization friendly, and flexible. Extensive experiments on image classification, object detection and segmentation benchmarks with both convolutional networks (ResNet and MobileNetV2) and vision transformers were consistently better than previous state-of-the-art quantization aware training methods.

\section{Acknowledgments}
This research was partly supported by the National Natural Science Foundation of China under Grant 62276123 and Grant 61921006.

\bibliography{aaai23}

\end{document}